\def\year{2019}\relax
\documentclass[letterpaper]{article} 
\usepackage{aaai19}  
\usepackage{times}  
\usepackage{helvet}  
\usepackage{courier}  
\usepackage{url}  
\usepackage{graphicx}  
\usepackage{color}
\usepackage{multirow}
\usepackage{booktabs} 
\definecolor{darkgreen}{rgb}{0,0.6,0.2}

\newcommand{\tabincell}[2]{\begin{tabular}{@{}#1@{}}#2\end{tabular}}

\begin{document}
%
\title{End-to-End Knowledge-Routed Relational Dialogue System for\\ Automatic Diagnosis}
\author{Lin Xu$^{1\dagger}$,\ Qixian Zhou$^{1\dagger}$,\ Ke Gong$^1$,\ Xiaodan Liang$^{1}$\thanks{Lin Xu and Qixian Zhou contribute equally to this work and share first-authorship. Corresponding author is Xiaodan Liang.},\ Jianheng Tang$^3$,\ Liang Lin$^{1,2}$  \\
$^1$Sun Yat-Sen University, $^2$Dark Matter AI Inc. , $^3$Soochow University\\
\tt\small{cathyxl2016@gmail.com,\ qixianzhou.mail@gmail.com,\ gongk3@mail2.sysu.edu.cn} \\
\tt\small{xdliang328@gmail.com,\ sqrt3tjh@gmail.com,\ linliang@ieee.org}
}
\maketitle
\begin{abstract}
Beyond current conversational chatbots or task-oriented dialogue systems that have attracted increasing attention, we move forward to develop a dialogue system for automatic medical diagnosis that converses with patients to collect additional symptoms beyond their self-reports and automatically makes a diagnosis. Besides the challenges for conversational dialogue systems (e.g. topic transition coherency and question understanding), automatic medical diagnosis further poses more critical requirements for the dialogue rationality in the context of medical knowledge and symptom-disease relations. Existing dialogue systems~\cite{madotto2018mem2seq,wei2018task,li2017end} mostly rely on data-driven learning and cannot be able to encode extra expert knowledge graph. In this work, we propose an End-to-End Knowledge-routed Relational Dialogue System (KR-DS) that seamlessly incorporates rich medical knowledge graph into the topic transition in dialogue management, and makes it cooperative with natural language understanding and natural language generation. A novel Knowledge-routed Deep Q-network (KR-DQN) is introduced to manage topic transitions, which integrates a relational refinement branch for encoding relations among different symptoms and symptom-disease pairs, and a knowledge-routed graph branch for topic decision-making. Extensive experiments on a public medical dialogue dataset show our KR-DS significantly beats state-of-the-art methods (by more than 8\% in diagnosis accuracy). We further show the superiority of our KR-DS on a newly collected medical dialogue system dataset, which is more challenging retaining original self-reports and conversational data between patients and doctors.
\end{abstract}

\section{Introduction}
\begin{table}
\small
\tabcolsep 0.02in
\begin{tabular}{lll}
\toprule[0.7pt]
  \multicolumn{3}{l}{\textbf{Self-report}} \\
  \hline
  \multicolumn{3}{l}{P: Baby \underline{\textbf{vomited}} last night. What is the reason?} \\
  \hline
  \multicolumn{3}{l}{\textbf{Baseline Conversation}} \\
  \hline                           
  A: & Dose baby have \underline{diarrhea}?    & P:  Maybe not.\\
  A: & Does baby have \underline{running nose}?  & P:  Not sure.\\
  A: & Does baby have \underline{sputum}?      & P:  Not sure.\\
  A: & Baby have \underline{sputum}, doesn't he?   & P:  Not sure.\\
  A: & Your baby may have \textit{children dyspepsia}. & \\
  \hline
  \multicolumn{3}{l}{\textbf{KR-DS Conversation}} \\
  \hline
  A: & Does baby have \underline{fever}?   & P:  Not sure.\\
  A: & Baby \underline{doesn't want to eat (\textbf{anorexia})}, is it? & P: Right!\\
  A: & Does he have \underline{\textbf{bloating}}?   &  P: Baby does have.\\
  A: & Your baby may have \textit{children dyspepsia}.  & \\
\toprule[0.7pt]
\end{tabular}
\caption{A conversation example between the patient (P) and the agent (A). We underline all symptoms and highlight the symptoms related to the diagnosed disease (italic). Superior to the baseline method~\cite{wei2018task} that asks irrelevant and repeated symptoms, our KR-DS generates more reasonable results thanks to the relational refinement and guidance of knowledge-routed graph, as presented in Fig.~\ref{fig:dm_framework}.  }
\label{table:muzhi_res}
\end{table}

Task-oriented dialogue aims to achieve specific task through interactions between the system and users in natural language, which is gaining research interest in the different application domain, including movie booking~\cite{lipton2017bbq}, restaurant booking~\cite{wen2017network}, online shopping~\cite{yan2017building}, and technical support~\cite{lowe2015ubuntu}. In the medical domain, A dialogue system for medical diagnosis converses with patients to obtain additional symptoms and make a diagnosis automatically, which has significant potential to simplify the diagnostic procedure and reduce the cost of collecting information from patients~\cite{tang2016inquire}. Besides, patient condition reports and preliminary diagnosis reports generated by the dialogue system may assist doctors to make a diagnosis more efficiently.

However, the dialogue system for medical diagnosis poses stringent requirements not only on the dialogue rationality in the context of medical knowledge but the comprehension of symptom-disease relations. The symptoms which dialogue system inquiries should be related with underlying disease and consistent with medical knowledge. Current task-oriented dialogue systems~\cite{lei2018sequicity,lukin2018scoutbot,bordes2016learning} highly rely on the complex belief tracker~\cite{wen2017network,mrkvsic2016neural} and pure data-driven learning, which are unable to apply to automatic diagnosis directly for the lack of considering medical knowledge. A very recent work~\cite{wei2018task} made the first move to build a dialogue system for automatic diagnosis, which cast dialogue systems as Markov Decision Process and trained the dialogue policy via reinforcement learning. Nevertheless, this work only exploits Deep Q-network (DQN) via data-driven learning to manage topic transitions (deciding which symptoms should be asked), whose results are intricate and repeated, as shown in Table~\ref{table:muzhi_res}. Moreover, this work only targets dialogue management for dialogue state tracking and policy learning, utilizing extra template-based models for natural language processing, which fails to match the real-world automatic diagnosis scenarios well. 

\begin{figure*}[t]
\centering
  \includegraphics[width=0.7\linewidth]{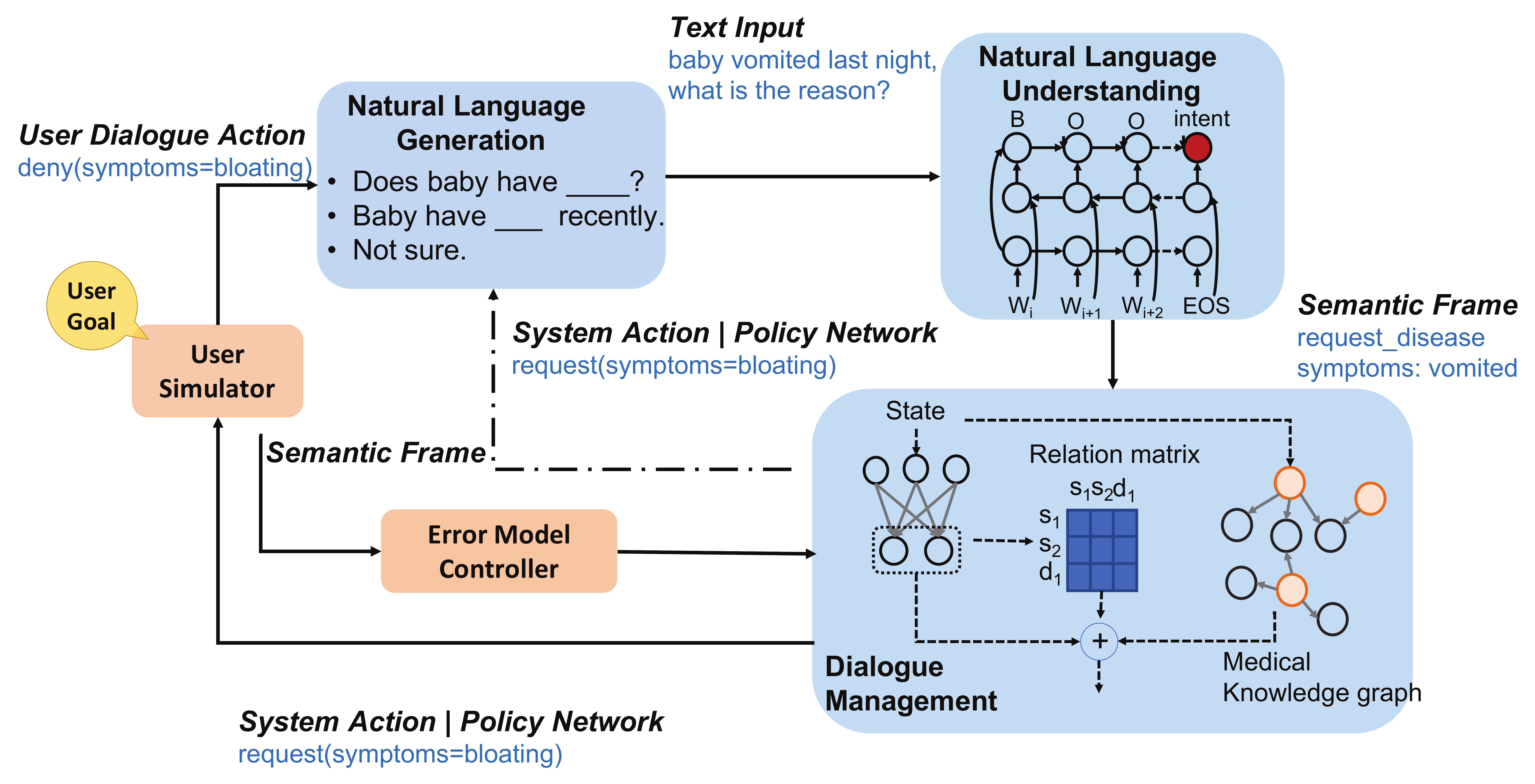}
\caption{Framework of our end-to-end Knowledge-routed Relational Dialogue System (KR-DS), A Bi-LSTM based Natural Language Understanding is employed to parse input utterances and generate semantic frames which are further fed into DM module to generate system actions. Dialogue Management manages the system actions containing a basic DQN branch, a relational refinement branch, and a medical knowledge-routed graph branch. A User Simulator with a user goal (consisting of symptoms of patients) to interact with agent and give rewards. A template-based NLG is used to generate natural language for user simulator and dialogue manager based on actions. 
}
\label{fig:system_framework}
\end{figure*}

To address the above challenges, we propose a complete end-to-end Knowledge-routed Relational Dialogue System (KR-DS) for automatic diagnosis, that seamlessly incorporates rich medical knowledge graph and symptom-disease relations into the topic transition in Dialogue Management (DM), and makes it cooperative with Natural Language Understanding (NLU) and Natural Language Generation (NLG), as shown in Fig.~\ref{fig:system_framework}. 

In general, doctors determine a diagnosis based on medical knowledge and diagnosis experience. Inspired by that, we designed a novel Knowledge-routed Relational Deep Q-network (KR-DQN) for dialogue management, which integrates a knowledge-routed graph branch and a relational refinement branch, taking full advantage of the medical knowledge and historical diagnostic cases (doctor experience). The relational refinement branch automatic learns relations among different symptoms and symptom-disease pairs from historical diagnostic data to refine rough results generated from a basic DQN. The knowledge-routed graph branch helps policy decision via a well-designed medical knowledge graph routing prior information based on conditional probabilities. Therefore, these two branches can contribute to generating more reasonable decision results from knowledge guiding and relation encoding, as shown in Table \ref{table:muzhi_res}. 

Moreover, existed dataset~\cite{wei2018task} fails to support our end-to-end dialogue system training as it only contains user goals (extracted and normalized symptoms and disease) instead of complete conversation data. Therefore, we build a new dataset collected from an online medical forum by extracting symptoms and diseases from patients’ self-reports and conversations between patients and doctors. Our dataset reserves the original self-reports and interaction utterances between doctors and patients to train the NLU component with real dialogue data, which matches the realistic clinic scenarios much better.

Our contributions are summarized in the following aspects. 1) we propose an End-to-End Knowledge-routed Relational Dialogue System (KR-DS) that seamlessly incorporates medical knowledge graph into the topic transition in dialogue management, and makes it cooperative with natural language understanding and natural language generation. 2) A novel Knowledge-routed Deep Q-network (KR-DQN) is introduced to manage topic transitions, which integrates a relational refinement branch for encoding relations among different symptoms and symptom-disease pairs, and a knowledge-routed graph branch for topic decision-making. 3) We construct a new challenging end-to-end medical dialogue system dataset, which retains the original self-reports and the conversational data between patients and doctors. 4) Extensive experiments on two medical dialogue system datasets show the superiority of our KR-DS, which significantly beats state-of-the-art methods by more than 8\% in diagnostic accuracy.

\section{Related Work}
The successes of RNNs architecture~\cite{wen2017network,serban2016building,zhao2017generative} motivated investigation in dialogue systems due to its ability to create a latent representation, avoiding the need for artificial state labels. Sequence-to-sequence models have also been used in task-oriented dialogue systems~\cite{sutskever2014sequence,vinyals2015neural,eric2017copy,madotto2018mem2seq}.~\cite{zhao2017generative} proposed framework enabling encoder-decoder models to accomplish slot-value independent decision-making and interact with external databases.~\cite{chen2018hierarchical} proposed a hierarchical memory network by adding the hierarchical structure and the variational memory network into a neural encoder-decoder network. Although these architectures have better language modeling ability, they do not work well in knowledge retrieval or knowledge reasoning, where the model inherently tends to generate short and general responses in spite of different inputs.

Another research line comes from the utilizing of knowledge bases.~\cite{young2017augmenting} investigated the impact of providing commonsense knowledge about the concepts covered in the dialogue.~\cite{liu2018knowledge} proposed a neural knowledge diffusion model to introduce knowledge into dialogue generation.~\cite{eric2017key} noticed that neural task-oriented dialogue systems often struggle to smoothly interface with a knowledge base and they addressed the problem by augmenting the end-to-end structure with a key-value retrieval mechanism.

In addition, there are two works much related to our concerns where deep reinforcement learning is applied for automatic diagnosis~\cite{tang2016inquire,wei2018task}. However, they only targeted on dialogue management for dialogue state tracking and policy learning. Moreover, their data used is simulated or simplified that cannot reflect the situation of the real diagnosis. In our work, we perform not only relation modeling to attend correlated symptoms and diseases but also graph reasoning to guide policy learning by prior medical knowledge. We further introduce an end-to-end task-oriented dialogue system for automatic diagnosis and a new medical dialogue system dataset, which pushes the research boundary of automatic diagnosis to match real-world scenarios much better.

\section{Proposed Method}
Our End-to-End Knowledge-routed Relational Dialogue System (KR-DS) is illustrated in Fig.~\ref{fig:system_framework}. As a task-oriented dialogue system, it contains Natural Language Understanding (NLU), Dialogue Management (DM) and Natural Language Generation (NLG). NLU recognizes user intent and slot values from utterances. Then DM executes topic transition according to current dialogue state. The agent in DM would learn to request symptoms to proceed diagnosis task and inform disease to make a diagnosis. Given the predicted system actions, natural language sentences are generated by a template-base NLG. Additionally, to train the whole system in an end-to-end way via reinforcement learning, a user simulator is added to execute the conversation exchange conditioned on the generated user goal.

\subsection{Natural Language Understanding}
In our task, NLU is mainly used to classify intent and fill slots for the Chinese language as our dataset was collected from a Chinese website. We use a public Chinese segmentation tool and add medical terms into the custom dictionary to improve accuracy. 
Given a word sequence, NLU classifies intent and fill in a set of slots to form a semantic frame. As illustrated in Fig.~\ref{fig:system_framework}, semantic frames are structured data that contain the user's intent and slots. In particular, six types of intents are considered in our automatically diagnosis dialogue system. For a user, there are four types of intents, including \emph{request+disease}, \emph{confirm+symptom}, \emph{deny+symptom} and \emph{not-sure+symptom}. We apply the popular BIO format ~\cite{hakkani2016multi} to label each word tag in the sentence. 

Following \cite{hakkani2016multi}, given a word sequence, we apply a Bi-LSTM to recognize BIO tags of each word and simultaneously classify intent of this sentence. With tag labeling done we fill slots based on dialogue contextual information and medical term normalization. Symptoms and diseases are normalized to medical terms defined by the annotators. With regard to context understanding, we maintain a rule-based dialogue state tracker, which stores the status of symptoms. We represent slots of the current semantic frame as a fixed-length symptom vector (each bit -1, 1, -2, 0) and plus it to last symptom vector to get a new symptom vector. If a symptom is requested by an agent and happens not to appear in user’s answer, this method could also fill slot by referring to the recorded request slot.

As symptoms and intents are labeled in our dataset, we use supervised learning to train Bi-directional LSTM model. Moreover, after pre-training, NLU can be jointly trained with other parts in our KR-DS through reinforcement learning.

\subsection{Policy Learning with KR-DQN}

\textbf{Overview.} We design DM in a reinforcement learning framework. Dialogue Manager is an agent and interacts with the environment (user simulator) via the dialogue policy. The optimized dialogue policy selects the best action that maximizes the future reward, which suitably solves the problem of our diagnosis dialogue system for its large selection spaces, sequential decision and success/failure end. As shown below, we first describe the basic elements of reinforcement learn- ing in our task.
Suppose we have M diseases, N symptoms. Four types of agent action include\emph{inform+disease}, \emph{request+symptom}, \emph{thanks} and \emph{closing}. Therefore agent action space size D is denoted as $D =num\underline{\hspace{0.5em}}greeting+M+N$. User actions are \emph{request+disease}, \emph{confirm/deny/not-sure+symptom} and \emph{closing}. There are also four types of symptom states, \emph{positive}, \emph{negative}, \emph{not-sure} and \emph{not mentioned}, represented by 1, - 1, -2, 0 in symptom vectors respectively. At each turn, dialogue state $s_t$ contains the previous action of both user and agent, known symptoms representation and current turn information.

Rewards are crucial for policy learning. In our task, we encourage the agent to make the right diagnosis and penalize the wrong diagnosis. Moreover, brief dialogue and precise symptoms request are encouraged. Consequently, we design a reward as follows, +44 for successful diagnosis and -22 for failure. For each turn, we only apply -1 penalty for failing to hit the existed symptom request on each turn. As for the comparison of different reward, we demonstrate several experimental results in ablation parts. Finally, dialogue policy $\pi$ describes the behaviors of an agent. It takes state st as input and outputs the probability distribution over all possible actions $\pi(a_t|s_t)$.

\begin{figure}[t]
\centering
  \includegraphics[width=1\linewidth]{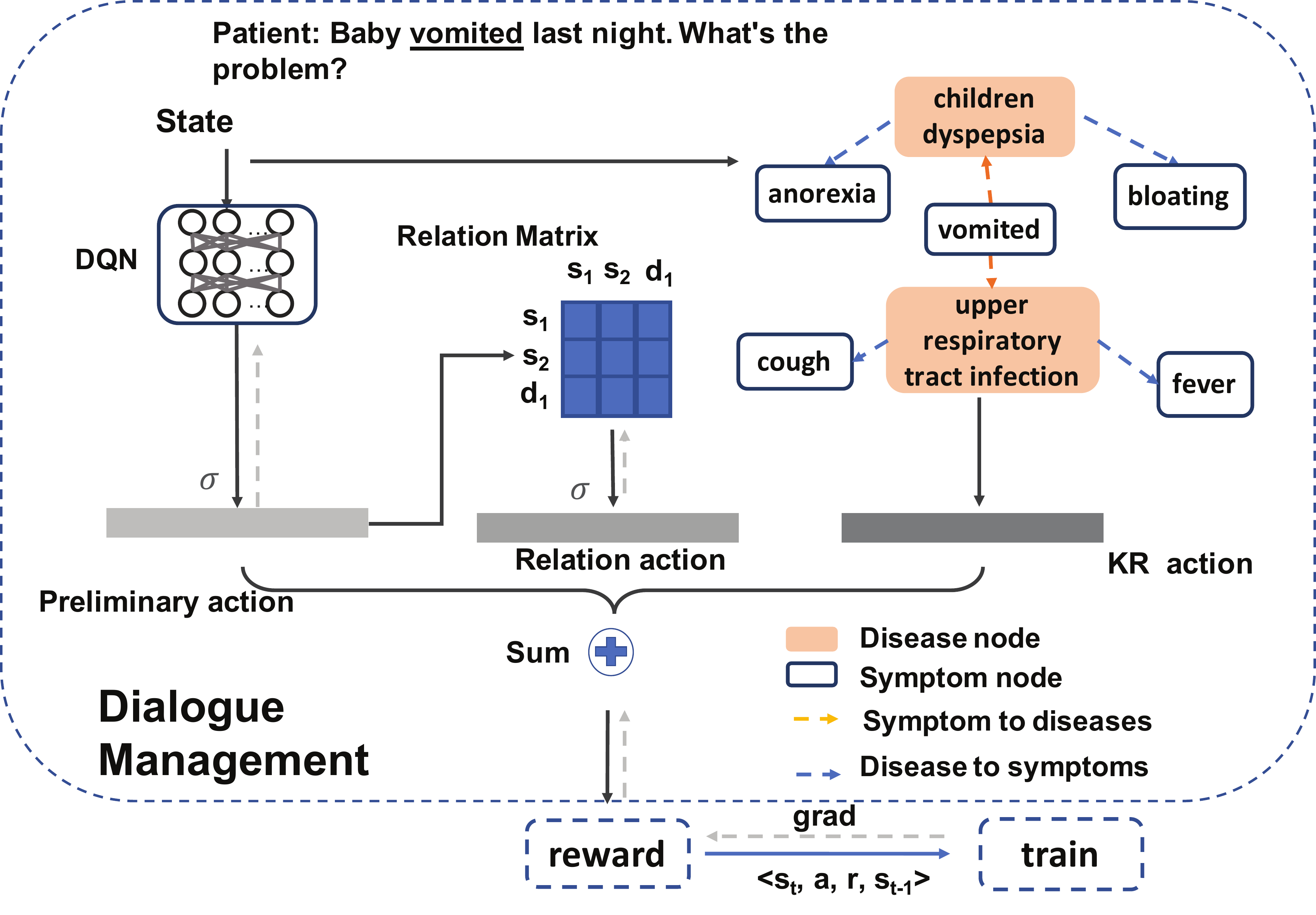}
\caption{Illustration of Knowledge-routed Deep Q-network (KR-DQN) for DM. The basic DQN branch generates rough action results. The relational branch encodes relations among actions to refine the results. The knowledge-routed graph branch induces medical knowledge for graph reasoning and conducts a rule-based decision to enhance the action results. The three branches can be jointly trained through reinforcement learning.}
\label{fig:dm_framework}
\end{figure}

DQN is a common policy network in many problems, including playing game~\cite{mnih2015human}, video captioning ~\cite{wang2018video}. Beyond the DQN with simple Multi-Layer Perceptron(MLP), we propose a novel Knowledge-routed Relational DQN (KR-DQN) by considering prior medical knowledge and modeling relations between actions to generate reasonable actions, illustrated in Fig.~\ref{fig:dm_framework}.

\textbf{Basic DQN Branch.}
We first utilize a basic DQN branch to generate a rough action result, as formulated in Eq.~(\ref{eq:basic_dqn}) The MLP takes state $s_t$ as input and outputs a rough action result $a^r_t \in R^D$. The structure of MLP is a simple neural network with a hidden layer. The activation function of the hidden layer is rectified linear unit. 
\begin{equation}
a^r_t = MLP(s_t).
\label{eq:basic_dqn}
\end{equation}

\textbf{Relational Refinement Branch.}
A relation module can affect an individual element (e.g. a symptom or a disease) by aggregating information from a set of other elements (e.g. other symptoms and diseases). As the relation weights are automatically learned driven by the task goal, the relation module can model dependency between the elements. So we design a relational refinement module by introducing a relation matrix $R \in R^{D \times D}$ which represents dependency among all actions. The initially predicted action $a^r_t$ subsequently multiplies this learnable relation matrix $R$ to get refined action result $a^f_t \in R^D$, which is written as Eq.~(\ref{eq:relation_refinement}). The relation matrix is asymmetric, representing a directed weighted graph. Each column of the relation matrix sums to one and each elements of refined action vector $a^f_t$ is a weighted sum of the initially predicted action $a^r_t$, where weights denote the dependency between the elements.
\begin{equation}
a^f_t = a^r_t {\cdot} R.
\label{eq:relation_refinement}
\end{equation}
The relation matrix $R$ is initialized with conditional probabilities from dataset statistics. Each entry $R_{ij}$ denotes the probability of the unit $x_j$ conditional on the unit $x_i$. The relation matrix learns to capture the dependency among actions through back-propagation. Our experiments show that this initialization method is better than random initialization as the prior knowledge can guide relation matrix learning. 

\textbf{Knowledge-routed Graph Branch.}
When receiving patients' self-reports, doctors first grasp a general understanding of the patients with several possible candidate diseases. Subsequently, by asking for significant symptoms of these candidate diseases, doctors exclude other candidate diseases until confirms a diagnosis. Inspired by this, we design a knowledge-routed graph module to simulate the thinking process of a doctor.

We first calculate the conditional probabilities between diseases and symptoms as directed medical knowledge-routed graph weights, in which there are two types of nodes (diseases and symptoms) as shown in Fig.~\ref{fig:dm_framework}. Edges only exist between a disease and a symptom. Each edge has two weights, which are conditional probabilities from diseases to symptoms, noted as $P(dis|sym) \in R^{M \times N}$, and conditional probabilities from symptoms to diseases, written as $P(sym|dis) \in R^{N \times M}$. 

During communication with patients, doctors may have several candidate diseases. We represent candidate diseases probability as disease probabilities corresponding to observed symptoms. Symptom prior probabilities $P_{prior}(sym) \in R^N$ are calculated through the following rules. For the mentioned symptoms, positive symptoms is set to 1 while negative symptoms is set to -1 to discourage its related diseases. Other symptom(not sure or not mention) probabilities are set to its prior probabilities which are calculated from the dataset. Then these symptom probabilities $P_{prior}(sym)$ multiply conditional probabilities $P(dis|sym)$ to get disease probability $P(dis)$, which is formulated as:
\begin{equation}
P(dis) = P(dis|sym) {\cdot} P_{prior}(sym).
\end{equation}
Considering candidate diseases, doctors often inquire some notable symptoms to confirm diagnosis according to their medical knowledge. Likewise, with disease probabilities $P(dis)$, symptom probabilities $P(sym)$ is obtained by matrix multiplication between diseases probabilities $P(dis)$ and conditional probabilities matrix $P(sym|dis)$:
\begin{equation}
P(sym) = P(sym|dis) {\cdot} P(dis).
\end{equation}
We concatenate disease probabilities $P(dis) \in R^M$ and symptom probabilities $P(sym) \in R^N$ padded with zeros for greeting actions to get knowledge-routed action probabilities $a^k_t \in D$.

With three action vectors $a^r_t$, $a^f_t$ and $a^k_t$, we first apply $sigmoid$ activation function to action vector $a^r_t$ and $a^f_t$ to obtain action probabilities. Subsequently, we sum these three results as predicted action distributions $a_t$ of KR-DQN under the current state $s_t$.
\begin{equation}
a_t = sigmoid(a^r_t) + sigmoid(a^f_t) + a^k_t
\end{equation}
In order to prevent repeated request, we add symptoms filter to KR-DQN outputs. All components are trained with a well-designed reward (described at the beginning of this section) to encourage KR-DQN learn how to request effective symptoms and make a right diagnosis. 

\subsection{User Simulator}

\begin{figure}[t]
\centering
  \includegraphics[width=1\linewidth]{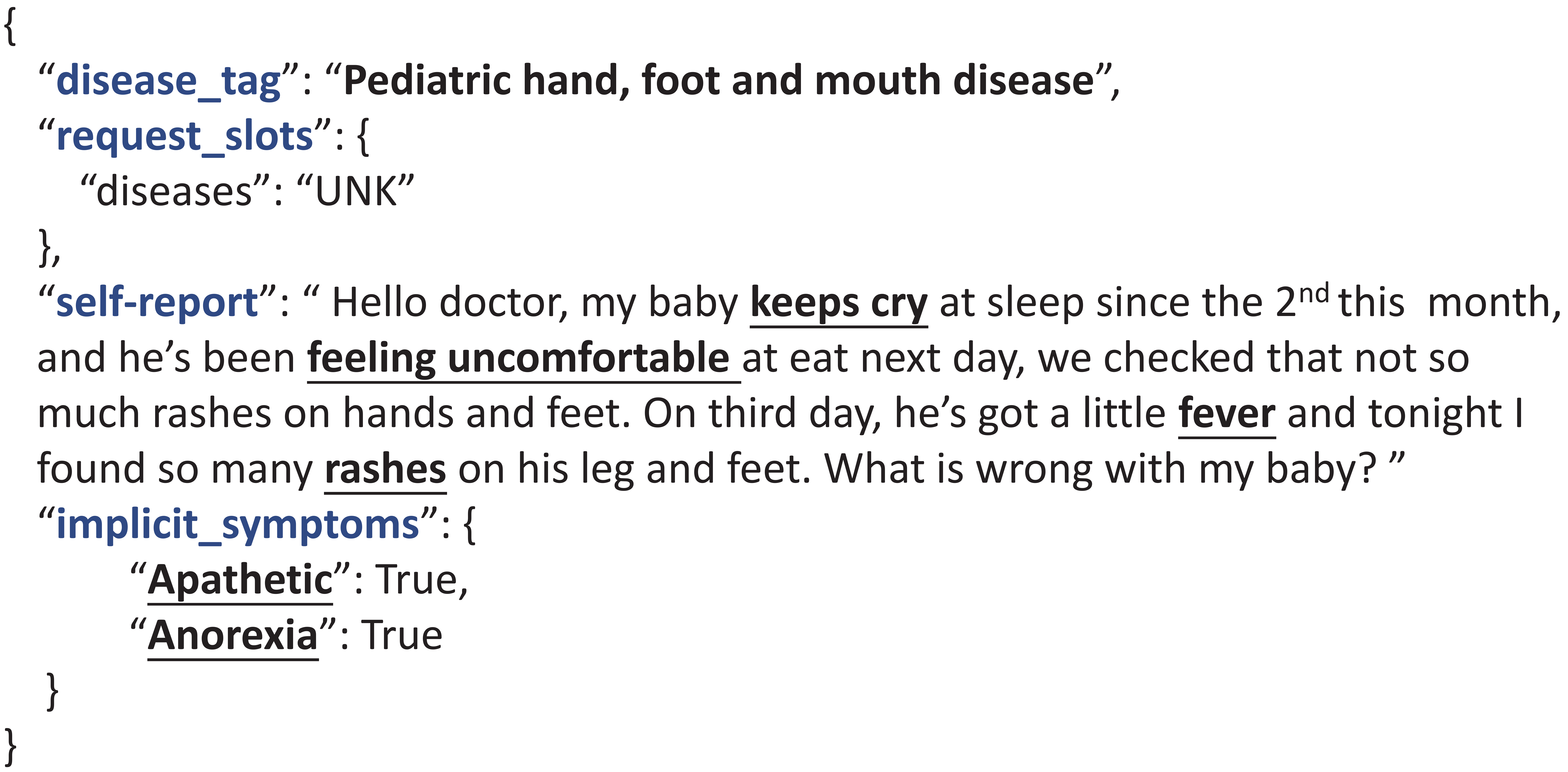}
\caption{An example of a user goal.}
\label{fig:user_goal}
\end{figure}

In order to train our end-to-end dialogue system, we apply a user simulator to sample user goals from the experimental dataset for automatically and naturally interacting with the dialogue system. Following ~\cite{schatzmann2009hidden}, our user simulator maintains a user goal G. As shown in Fig.~\ref{fig:user_goal}, a user goal generally consists of four parts: \textit{disease tag} for the disease that the user suffers, \textit{self-report} for the original self-reports from patients, \textit{implicit symptoms} for symptoms talked about between the patient and the doctor, and \textit{request slots} for the disease slot that the user would request. When the agent requests a symptom during the course of the dialogue, the user will take one of the three actions including \textit{True} for the positive symptom, \textit{False} for the negative symptom, and \textit{Not sure} for the symptom that is not mentioned in the user goal. The dialogue session will be terminated as successful if the agent informs a correct disease. On the contrary, the dialogue process will fail if the agent makes the wrong diagnosis or the dialogue turn reaches the maximum turn \textit{T}.

\subsection{Natural Language Generation}
Given actions produced from Dialogue Management and User Simulator, a template-based natural language generation (template-NLG) is applied in our system to generate human-like sentences. As mentioned in NLU part, request and inform pairs are relatively simple. Previous dialogue systems~\cite{li2017end,lei2018sequicity} have many possible request/inform patterns but each one of them has only one template. Varying from them, we design 4 to 5 templates for each action to diversify dialogues. As for medical terms used in the dialogues, analogous to NLU, we choose daily expressions corresponding to specific symptoms and diseases from our collected medical term list.

\subsection{End-to-End Training With Deep Q-Learning}
Following~\cite{mnih2015human}, we employ Deep Q-Learning to train DM with fine-tuned NLU and template-base NLG. Two important DQN tricks~\cite{van2016deep}, target network usage and experience replay are applied in our system. 
We use $Q(s_t, a_t|\theta)$ to denote the the expected discounted sum of rewards, after taking a action $a_t$ under state $s_t$. Then according Bellman equation, the Q-value can be written into:
\begin{equation}
Q(s_t, a_t|\theta) = r_t + {\gamma}{\max}_{a_{t+1}}Q^{\ast}(s_{t+1}, a_{t+1}|{\theta}')
\end{equation}
${\theta}'$ is the parameters of target network obtained from previous episode. $\gamma$ is the discount rate. We use $\epsilon$-greedy exploration at training phase for effective action space exploration, selecting a random action in probability $\epsilon$. We store the agent’s experiences at each time-step in experience replay buffer, denoted as $e_t(s_t, a_t, r_t, s_{t+1})$. The buffer is flushed if current network performs better than all previous models. 

\section{Experiments}
\subsection{DX Medical Dialogue Dataset}
We build a newly DX dataset for medical dialogue system, reserving the original self-reports and interaction utterances between doctors and patients. We collected data from a Chinese online health-care community (dxy.com) where users asking doctors for medical diagnosis or professional medical advice.  
We annotate five types of diseases, including \textit{allergic rhinitis}, \textit{upper respiratory infection}, \textit{pneumonia}, \textit{children hand-foot-mouth disease}, and \textit{pediatric diarrhea}. We extract the symptoms that appear in self-reports and conversation and normalize them into 41 symptoms. Four annotators with medical background are invited to label the symptoms in both self-reports and raw conversations.Symptoms appearing in self-reports are regarded as explicit symptoms while the others are implicit symptoms. The diseases of each medical diagnosis conversation are labeled automatically by the website. There are 527 conversational data in total. 423 conversational data are selected as the training set 104 for testing. More detailed dataset statistics are shown in Table~\ref{table:dataset_statistics}. 

\begin{table}
\tabcolsep 0.08in 
\footnotesize
\begin{tabular}{l|c|c}
\toprule[0.8pt]
  \textbf{Disease} & \textbf{Quantity} & \textbf{Symptoms}\\
  \hline
  Allergic rhinitis & 102 & 24\\
  \tabincell{c}{Upper respiratory tract infection} & 122 & 23\\
  Pneumonia & 100 & 29\\
  \tabincell{c}{Children hand-foot-mouth disease} & 101 & 22\\
  Pediatric diarrhea & 102 & 33\\
\toprule[0.8pt]
\end{tabular}
\label{table:dataset_statistics}
\caption{Statistics of dialogues and symptoms for diseases.}
\end{table}

\subsection{Experimental Setup}
\textbf{Datasets}. ~\cite{wei2018task} constructed a dataset by collecting data from Baidu Muzhi Doctor website, denoted as MZ dataset in this paper. The MZ dataset contains 710 user goals and 66 symptoms, covering 4 types of diseases. 
As the MZ dataset only contains user goal data, we just train the DM model with user simulator and error model controller using the provided train/test sets. The slot error rate and intent error rate are both set at 5\%. We further evaluate our end-to-end framework on our DX dataset that reserves original conversation data. We selected 423 dialogues for training and conducted inference on another 104 dialogues.

\noindent\textbf{Evaluation Metrics}. Same as ~\cite{wei2018task}, we use the rate of making the right diagnosis as dialogue accuracy. We also employ a newly metric, namely matching rate, to evaluate the effectiveness of symptoms dialogue system requests. A successful matching means systems ask a symptom that exists in user implicit symptoms, otherwise, it is a failure matching. 

\noindent\textbf{Implementation Details}
We implement the system on Pytorch. To train the DQN composed of a two-layer neural network, the $\epsilon$ of $\epsilon$-greedy strategy is set to 0.1 for effective action space exploration and the $\gamma$ in Bellman equation is 0.9. The initial buffer size D is 10000 and the batch size is 32. The learning rate is 0.01. We choose SGD as the optimizer and 100 simulation dialogues will add to experience replay pool at each epoch training. Generally, we train the models for about 300 epochs. The source code will be released together with our DX dataset.

\begin{figure*}[t]
\centering
  \includegraphics[width=1\linewidth]{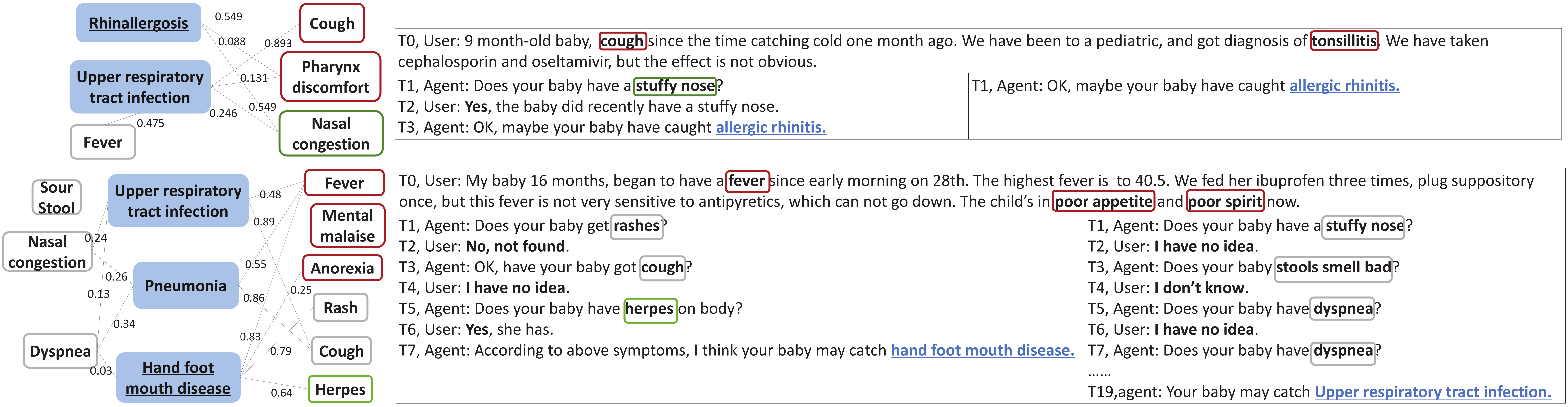}
\caption{Visualized conversation results on DX dataset. In each table, the first line is a self-report of patient. The results generated by our KR-DS are presented on the left column and the ones produced by the pure DQN method~\cite{wei2018task} on the right. We also present the knowledge graph related to the diagnosis process to the left of each table.
To highlight the symptoms and diseases, red boxes used for symptoms from self-report, green for \textit{True} symptom, gray for \textit{False} and \textit{not sure} symptoms in the user goal, blue for the diagnosed disease.}
\label{fig:visaul}
\end{figure*}

\subsection{Experimental Results}

\textbf{MZ dataset}. We only train Dialogue Management for the limitation of MZ dataset and compare our method with several baselines, as shown in Table \ref{table:muzhi_results}.``SVM-em" means the SVM model trained with just explicit symptoms and ``SVM-em\&im" is the SVM model using both explicit and implicit symptoms. Basic DQN is the proposed framework of~\cite{wei2018task}. The results of the above three baselines are provided by~\cite{wei2018task}. ``DQN+relation branch" means our proposed model without knowledge-routed graph branch and ``DQN+knowledge branch" is the model without relation refinement branch.
Observed from Table \ref{table:muzhi_results}, the performance of SVM-em\&im is higher than SVM-em, which indicates that the implicit symptoms would make a significant improvement. However, basic DQN~\cite{wei2018task} gets 6\% loss of accuracy compared to SVM-ex\&im because it fails to inquiry effective implicit symptoms. Notably, our KR-DS not only significantly beats basic DQN (8\%) but also outperforms SVM-ex\&im (2\%), which shows that our method can inquiry implicit symptoms effectively and make a precise diagnosis, thanks to knowledge-routed graph reasoning and relational refinement. 

\noindent\textbf{DX dataset}. We further evaluate the proposed end-to-end KR-DS through Deep Q-learning on our DX dataset. For comparison, we re-implement a baseline, which shares the identical NLU and NLG with KR-DS but employ a single Deep Q-network~\cite{wei2018task} as policy network. In addition, we apply a state-of-the-art end-to-end task-oriented dialogue system framework~\cite{lei2018sequicity} to our task (Sequicity), which uses belief spans to store constraints and requests (which are symptoms and diseases in this task) for state tracking. 
As shown in Table \ref{table:dxy_results}, our method outperforms basic DQN~\cite{wei2018task} and state-of-art seq-to-seq~\cite{lei2018sequicity} method in both task complete accuracy and symptom matching rate. Focusing on obtaining the largest positive reward, basic DQN often guesses the right disease results but inquiries some unreasonable and repeated symptoms during the dialogue due to no constraints for symptom and disease relation (prior knowledge). Our framework shows superiority not only in a higher accuracy but also higher matching rate, which indicates the symptoms acquired by KR-DS agent is more reasonable and as a consequence, it can make the more right diagnosis. Seq-to-seq frameworks (Sequicity) performs worse on this medical diagnosis task as they focus on the in-dialogue sentence transition while ignoring medical symptom connections to diagnosis.
\begin{table}\centering
\scriptsize
\tabcolsep 0.03in 
\begin{tabular}{c|c|c|c|c|c}
\toprule[0.8pt]
  Method & \tabincell{l}{Infantile\\diarrhea} & \tabincell{l}{Dyspepsia} & \tabincell{l}{Upper\\respiratory\\infection} & \tabincell{l}{Bronchitis} & Overall\\
\hline
  SVM-ex & 0.89 & 0.28 & 0.44 & 0.71 & 0.59\\
  SVM-ex\&im & 0.91 & 0.34 & \textbf{0.52} & 0.93 & 0.71\\
  Basic DQN~\cite{wei2018task} & - & - & - & - & 0.65\\
\hline
  DQN + relation branch* & 0.87 & 0.31 & 0.42 & 0.86 & 0.68 \\
  DQN + relation branch & 0.92 & 0.35 & 0.49 & 0.93 & 0.70 \\
  DQN + knowledge branch & 0.88 & 0.31 & 0.44 & 0.89 & 0.68 \\
  Our KR-DS & \textbf{0.96} & \textbf{0.39} & 0.50 & \textbf{0.97} & \textbf{0.73}\\
\toprule[0.8pt]
\end{tabular}
\caption{Performance comparison on the MZ dataset.}
\label{table:muzhi_results}
\end{table}

\begin{table}\centering
\footnotesize
\tabcolsep 0.05in 
\begin{tabular}{c|ccc}
\toprule[0.8pt]
  Method & Accuracy & Match rate & Ave turns\\
  \hline
  \tabincell{c}{Basic DQN ~\cite{wei2018task}} & 0.731 & 0.110 & 3.92\\
  \tabincell{c}{Sequicity ~\cite{lei2018sequicity}} & 0.285 & 0.246 & 3.40\\
  Our KR-DS & \textbf{0.740} & \textbf{0.267} & 3.36\\
\toprule[0.8pt]
\end{tabular}
\caption{Performance comparisons with the state-of-the-art methods on DX dataset.}
\label{table:dxy_results}
\end{table}

\begin{table}\centering
\tabcolsep 0.16in 
\footnotesize
\begin{tabular}{c|cccc}
\toprule[0.8pt]
  Reward & R1 & R2 & R1* & R2* \\
  \hline
  Accuracy & 0.697 & 0.725 & 0.718 & 0.739\\
\toprule[0.8pt]
\end{tabular}
\caption{Evaluation of reward magnitude on MZ dataset.}
\label{table:reward_results}
\end{table}

\subsection{Ablation Studies}
\textbf{Component analysis.} To verify the effects of the main components of our KR-DS, we further conducted a series of ablation studies on MZ dataset, as shown in Table \ref{table:muzhi_results}. Here we mainly target at the following components in our framework: knowledge-routed graph branch and relation refinement branch. As is shown in Table \ref{table:muzhi_results}, all these factors contribute to better performance of our method. Additionally, initializing relation matrix with conditional probability in the relational branch is better than random initialization (``DQN+relation*''), as the prior medical knowledge can guide the relation matrix learning.

\noindent\textbf{Reward evaluation.} Our reward is designed based on the maximum turn value L=22, 2*L for success and -L for failure and -1 for the penalty. -1 penalty will cause shorter dialogue turns by accumulating through the process of dialogue. We evaluate several reward functions considering the magnitude of reward by doing experiments as follows: we chose four group of rewards R1: +22, -11, -1; R2: +11, -6, -1; R1*: +22, -11, -0.5; R2*: +11, -6, -0.25 for success, failure and penalty in experiments, and got accuracy shown in Fig.~\ref{table:reward_results}. We found using smaller reward value achieves similar results with ours, but leading to a stable training process.
\subsection{Human Evaluation}
\begin{figure}[t]
\centering
  \includegraphics[width=1\linewidth]{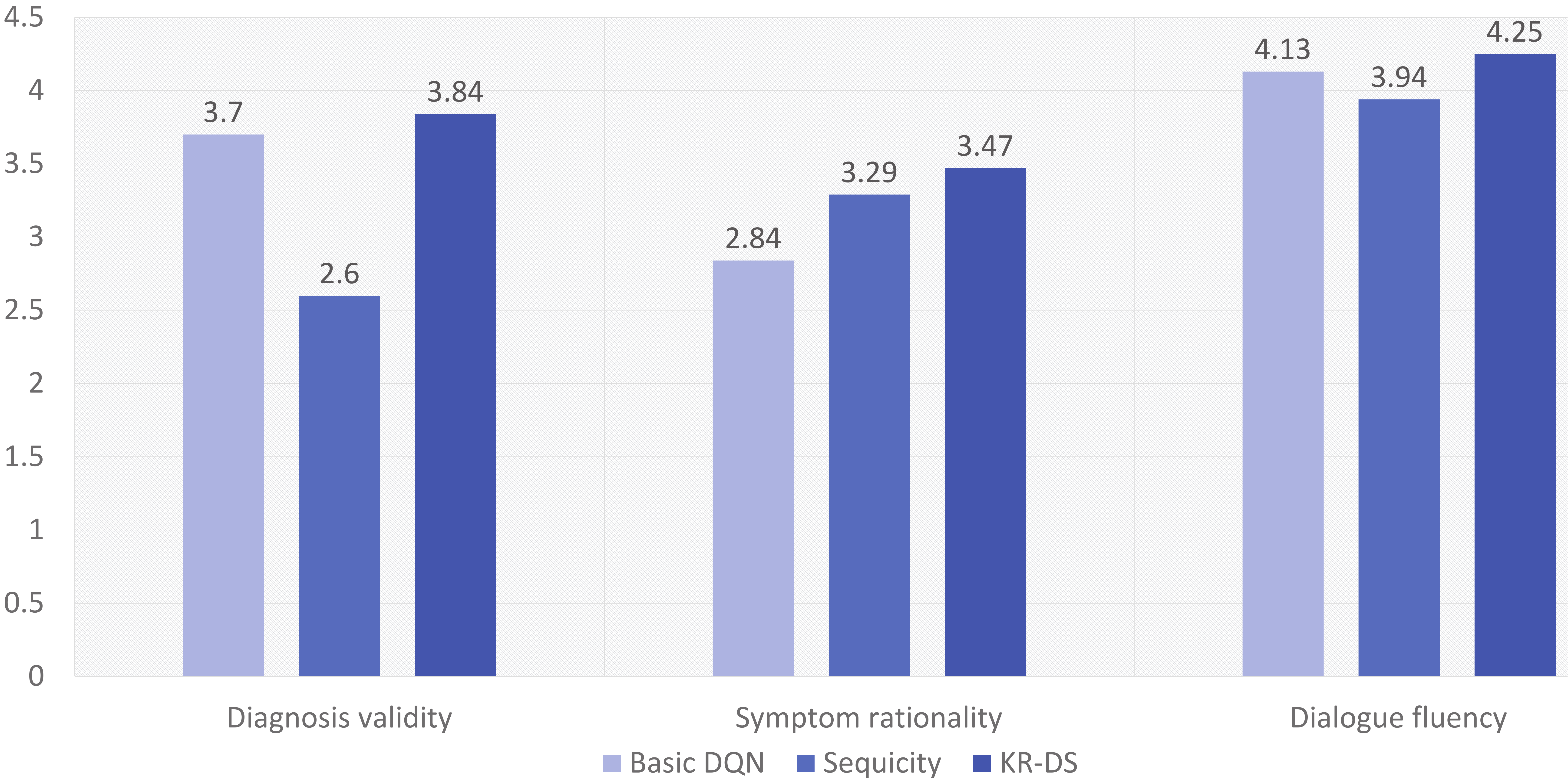}
\caption{Human evaluation of three methods.}
\label{fig:human_evaluation}
\end{figure}
As automated metrics is insufficient for evaluating a dialog system~\cite{liu2016not}, we invited three users with medical background to perform human evaluation based on three aspects: 1) diagnosis validity, 2) correctness of final diagnosis, symptom rationality and relevance, rationality of requested symptoms based on current dialogue status and medical knowledge, 3) dialogue fluency, dialogue fluency of communication. 104 dialogues in testing set are evaluated by three users. Users are given user goal (Fig.~\ref{fig:user_goal}) and converse with three dialogue systems, including Basic DQN~\cite{wei2018task}, Sequicity~\cite{lei2018sequicity} and our KR-DS. For each dialogue session, users are asked to giving a rating on a scale from 1 (worst) to 5 (best) on above three aspects. 

As shown in Fig.~\ref{fig:human_evaluation}, for diagnosis validity, our KR-DS has the higher average rating than other two methods. For symptom rationality and relevance, our method also outperforms other methods. Basic DQN obtains the lowest rating for requesting unrelated symptoms frequently. For dialogue fluency, our method also performs the best. 

\subsection{Qualitative Analysis}
For the more intuitive demonstration of our generated dialogues, two user goal results are displayed in Fig.~\ref{fig:visaul}. For the first example, self-report indicates two possible diseases, Rhinallergosis, and Upper respiratory tract infection according to the mentioned symptoms. The baseline gives the result directly, which may cause misdiagnosis. Contrarily, besides guidance by prior knowledge, our agent also considers the relations between symptoms and diseases, which tries to acquire a discriminatory symptom nasal congestion that is more relevant to Rhinallergosis and finally makes valid diagnosis. For the second result, the baseline method keeps asking symptoms which are not related to the correct diseases, like Sour Stool (no connection) and Nasal Congestion (with probability less than 0.3). Obviously, beneficial from the prior disease knowledge and symptom relations, our KR-DS asks Rash, Herpes, and Cough and diagnoses the correct Hand foot mouth disease.
\section{Conclusions}
In this work, we move forward to develop an End-to-End Knowledge-routed Relational Dialogue System (KR-DS) that enables dialogue management, natural language understanding, and natural language generation to cooperatively optimize via reinforcement learning. We propose a novel Knowledge-routed Deep Q-network (KR-DQN) upon a basic DQN to manage topic transitions, which further integrates a relational refinement branch for encoding relations among different symptoms and symptom-disease pairs, and a knowledge-routed graph branch for policy decision guided by medical knowledge. Additionally, we construct a new benchmark focusing on end-to-end medical dialogue systems, which retains the original self-reports and the conversational data between patients and doctors. Extensive experiments on two datasets show the superiority of our KR-DS, which generates the most precise and reasonable results.

\bibliographystyle{aaai}
\bibliography{egbib}

\end{document}